 \documentclass[smallabstract,smallcaptions]{dccpaper}

\usepackage{epsfig}
\usepackage{amsmath}
\usepackage{amssymb}
\usepackage{color}
\usepackage{url}
\usepackage{subcaption}
\usepackage{comment}
\usepackage{dirtytalk}
\usepackage{hyperref}
\usepackage{booktabs}
\newlength{\figurewidth}
\newlength{\smallfigurewidth}
\begin{document}

\title
{\large
\textbf{Fine-grained subjective visual quality assessment for high-fidelity compressed images}
}
 
\author{%
Michela Testolina$^{\ast,1}$, Mohsen Jenadeleh$^{\dag,1}$, Shima Mohammadi$^{\star}$, Shaolin Su$^{\dag}$,  
\\ João Ascenso$^{\star}$, Touradj Ebrahimi$^{\ast}$, Jon Sneyers$^{\diamond}$, and Dietmar Saupe$^{\dag}$\\[0.5em]
{\small\begin{minipage}{\linewidth}\begin{center}
\begin{tabular}{c}
$^{\ast}$ Multimedia Signal Processing Group, EPFL, Lausanne, Switzerland \\
\url{{michela.testolina, touradj.ebrahimi}@epfl.ch} 
\end{tabular}
\end{center}\end{minipage}}
\\ 
{\small\begin{minipage}{\linewidth}\begin{center}
\begin{tabular}{c}
$^{\dag}$ Multimedia Signal Processing Group, Universität Konstanz, Konstanz, Germany \\
\url{{mohsen.jenadeleh, shaolin.su, dietmar.saupe}@uni-konstanz.de} 
\end{tabular}
\end{center}\end{minipage}}
\\ 
{\small\begin{minipage}{\linewidth}\begin{center}
\begin{tabular}{c}
$^{\star}$ Instituto Superior Técnico, Lisbon, Portugal \\
\url{{shima.mohammadi, joao.ascenso}@lx.it.pt}\\
$^{\diamond}$ Media Technology Research Group, Cloudinary, Belgium, \url{jon@cloudinary.com}
\end{tabular}
\end{center}\end{minipage}}
\\[3.5em]
}
\footnotetext[1]{These authors contributed equally to this work.}

\maketitle
\thispagestyle{empty}

\begin{abstract}
Advances in image compression, storage, and display technologies have made high-quality images and videos widely accessible. 
At this level of quality, distinguishing between compressed and original content becomes difficult, highlighting the need for assessment methodologies that are sensitive to even the smallest visual quality differences. 
Conventional subjective visual quality assessments often use absolute category rating scales, ranging from ``excellent'' to ``bad''. While suitable for evaluating more pronounced distortions, these scales are inadequate for detecting subtle visual differences. 
The JPEG standardization project AIC is currently developing a subjective image quality assessment methodology for high-fidelity images. 
This paper presents the proposed assessment methods, a dataset of high-quality compressed images, and their corresponding crowdsourced visual quality ratings. It also outlines a data analysis approach that reconstructs quality scale values in just noticeable difference (JND) units.
The assessment method uses boosting techniques on visual stimuli to help observers detect compression artifacts more clearly. This is followed by a rescaling process that adjusts the boosted quality values back to the original perceptual scale. 
This reconstruction yields a fine-grained, high-precision quality scale in JND units, providing more informative results for practical applications. The dataset and code to reproduce the results will be available at \href{https://github.com/jpeg-aic/dataset-BTC-PTC-24}{https://github.com/jpeg-aic/dataset-BTC-PTC-24}.
\end{abstract}

\Section{Introduction}
The rapid advancements in image coding technologies and increasing demands for high-quality visual content have created a need for more sophisticated visual quality evaluation methodologies. Traditional subjective quality assessment techniques, like those presented in ITU-T Recommendation BT.500 \cite{BT.500} and reviewed in Part 1 of the JPEG AIC standard \cite{aic1}, are often effective for evaluating images with low and medium visual quality. However, when compared to quality scale reconstruction from pair comparisons, they lack precision \cite{mantiuk2012comparison}, and they fall short when adopted to evaluate the visual quality of high-fidelity contents, which requires distinguishing images with subtle variations in visual quality \cite{testolina2023performance}. For these reasons, the JPEG Committee launched a new activity in 2021, known as JPEG AIC-3 \cite{testolina2023assessment}, with the goal of a fine-grained quality assessment of compressed images with high-fidelity.

This paper presents and evaluates the anticipated JPEG AIC-3 subjective quality assessment methodology. In order to better distinguish and rank small-scale compression artifacts, two techniques are adopted. The first is artifact boosting, which emphasizes small differences between compressed and source images. The second is a triplet comparison in which observers compare two such compressed and boosted versions of the same source which is also displayed. From the collected responses that identify the stimulus with the perceived stronger distortion, a scaling procedure is used. It quantifies the perceived impairments on a linear scale, expressed in Just Noticeable Difference (JND) units. This scaling model assumes that the probability of correctly identifying the stimulus with a distortion that is 1 JND unit stronger than the one that it is compared with is 75\%. Since the goal is to estimate the perceived distortion of the original compressed images rather than the boosted ones, a rescaling procedure is applied. For this purpose, a limited number of triplet comparisons are carried out for a subset of the original compressed images, followed by a nonlinear regression procedure that aligns the two sets of reconstructed scale values. 

The proposed subjective quality assessment method was applied in a large-scale crowdsourcing campaign using five source images compressed with five recent and legacy compression methods yielding 250 decoded images (stimuli). About 440,000 triplet question responses were collected. The triplet responses, together with valuable auxiliary information such as subject IDs and response times, the resulting quality scale values, and the source code will be made publicly available.

\Section{Related work}
Subjective image quality assessment has been widely studied and standardized by organizations such as the International Telecommunication Union, with key recommendations found in ITU-T P.910 \cite{P.910} and  ITU-R BT.500 \cite{BT.500}. The two most commonly used methods for full-reference image quality assessment (IQA) are absolute category rating with hidden reference (ACR-HR) and 
double stimulus continuous quality scale (DSCQS). In ACR-HR, observers rate the quality of the source and the distorted image separately usually with five categories ranging. 
 In DSCQS observers are shown both the reference and the test images side by side or sequentially and asked to rate the quality of each image on a continuous scale.  The difference in their mean opinion scores gives the result, called DMOS.

The ISO/IEC 29170-2 standard, also known as JPEG AIC-2 \cite{aic2}, defines a flicker test to classify compressed images as either visually lossless or visually lossy. The boundary between lossy and lossless, as defined in this standard, can be understood as thresholding at the JND in the flicker test condition.  
Methods are also available to identify the distortion level or encoding parameter that produces a compressed stimulus at the JND threshold. In lab studies of \cite{lin2015experimental} and \cite{videoset}, comparisons between sources and distorted images or videos are used with binary search algorithms to estimate the JND thresholds per observer. Crowdsourced JND assessments using a slider-based interactive selection of the JND with the flicker test have shown promise for scalable subjective assessments of image quality \cite{lin2022large}. However, all of the above JND-based procedures can only distinguish between two stimuli if one of the encoded images is visually lossless (i.e., below 1 JND) and the other is not. 

The flicker test can be considered as a kind of boosting method to enhance observers' sensitivity in visual quality assessment, particularly in the high-quality range. Recent work has proposed zooming and artifact amplification as additional elements of boosting \cite{men2021subjective}. Finally, the triple stimulus boosted pairwise comparison (TSBPC), proposed in \cite{CID22}, follows a similar procedure and was adopted to create a large-scale dataset for medium to visually lossless quality image compression. Elements from these methods have been incorporated into the work presented here.

\Section{Proposed JPEG AIC-3 test methodologies}

\SubSection{Boosted triplet comparison (BTC)}
This methodology adopts boosting techniques to allow easier identification of subtle artifacts inspired by the work of \cite{men2021subjective}. The following boosting techniques are considered:
\begin{itemize}
    \item {Zooming}: Images are first cropped to half the size in both dimensions and then upscaled to the initial size using Lanczos resampling.  Alternatively, the size of the original images can be chosen to be small enough to allow zooming without cropping.
	\item {Artefact amplification}: The pixel-wise difference between the original and distorted stimuli are linearly scaled in the three color channels separately with an amplification factor of 2.
	\item {Flicker}: The test images are temporarily interleaved with the reference image at a change rate of 10\,Hz, i.e., in each cycle, the original image is shown for 100\,ms and then the distorted image is also displayed for 100\,ms. 
\end{itemize}

\begin{figure*}
    \centering
    \begin{subfigure}[b]{0.48\textwidth}
         \centering
         \includegraphics[height=4.6cm]{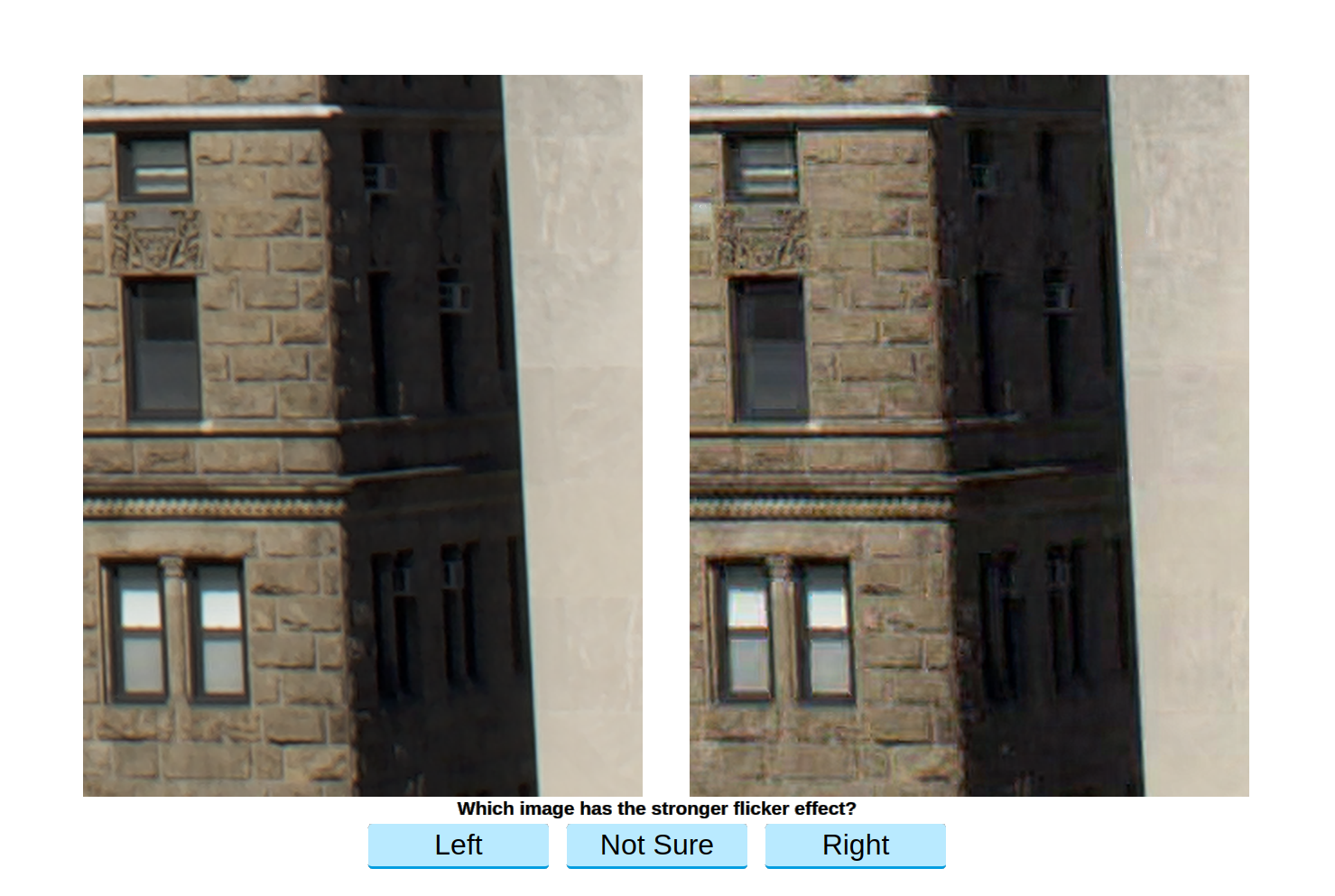}
         \caption{BTC}
         \label{fig_btc}
     \end{subfigure}
     \begin{subfigure}[b]{0.47\textwidth}
         \centering
         \includegraphics[height=4.8cm]{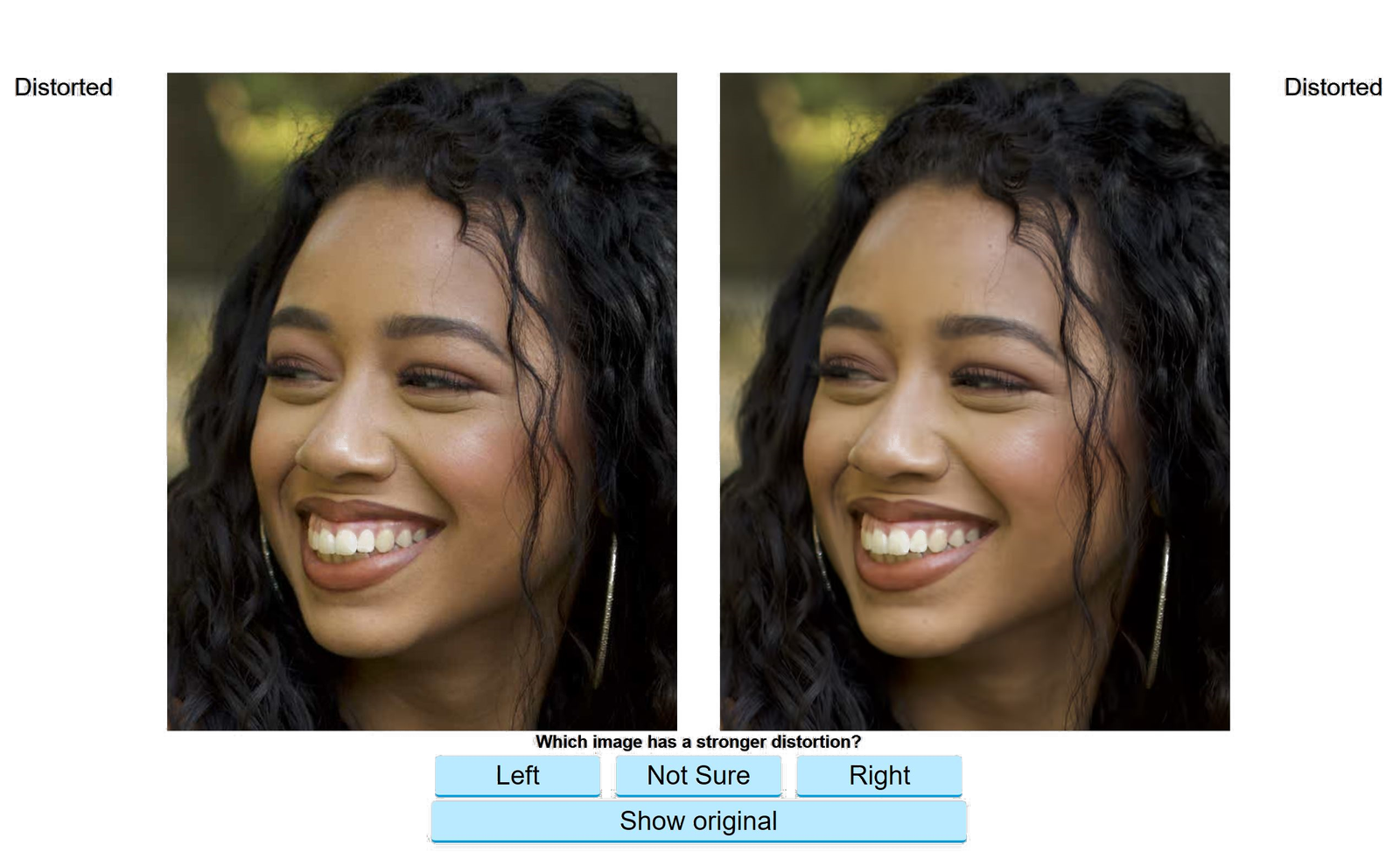}
         \caption{PTC}
         \label{fig_ptc}
     \end{subfigure}
    \caption{Sketch of the interface of the BTC and PTC experiment.}
    \label{fig_interface}
\end{figure*}

Triplets are denoted as $(I_i,I_0,I_k)$, where $I_i$ and $I_k$ are two compressed versions of the source image $I_0$. The two test stimuli $I_i$ and $I_k$ are displayed side-by-side, each alternating with the source to create a flicker on both sides (Figure~\ref{fig_btc}). For each such triplet, observers are asked to identify the stimulus with the strongest flicker effect, giving an answer ``Left”, ``Right”, or ``Not Sure”.  In \cite{jenadeleh2023relaxed}, it was shown that providing an indecision response option reduces mental load while maintaining the homogeneity of psychometric functions. Triplets are shown for 8 seconds and then blanked for 3 seconds. Anytime during the 11 seconds, subjects can select their answer, after which the next triplet is shown. The sequence of triplet questions is randomized and different for each observer.

\SubSection{Plain Triplet Comparison (PTC)}
PTC differs from BTC in a few key aspects. Most importantly, the decoded images are left untouched and used in place of the boosted versions. Source images are shown \textit{in-place} with the test images, but the flicker is replaced by a toggle button that the observer holds to switch between the decoded and the source image. A label indicates whether the currently displayed images are distorted or the source. Before submitting their answers, subjects are required to toggle at least once to make sure the original image is shown. The time window for answering is 30 seconds, and the maximal toggle frequency is limited to 2 Hz. The in-place presentation of PTC differences between a compressed image and the source image appear at the same locations on the display, thereby reducing eye movement and short-term memory needed for their detection, compared to side-by-side presentation. Still, the quality scale derived from PTC uses the plain decoded images and is regarded as the one that must be estimated at the end by the methodology.  

\Section{Experimental setup}

\begin{figure*}
    \centering
    \begin{subfigure}[b]{0.16\textwidth}
         \centering
         \includegraphics[width=\textwidth]{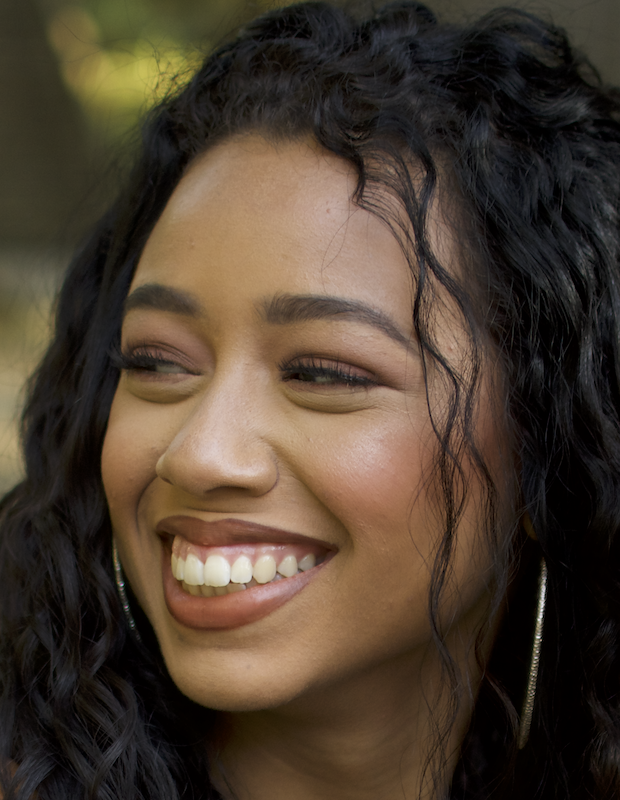}
         \caption{SRC 00002} 
     \end{subfigure}
     \begin{subfigure}[b]{0.16\textwidth}
         \centering
         \includegraphics[width=\textwidth]{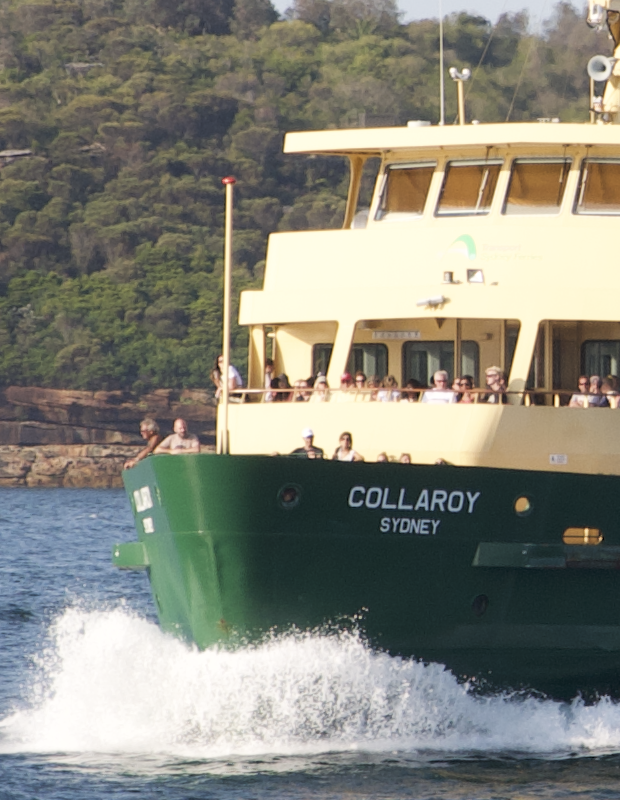}
         \caption{SRC 00006} 
     \end{subfigure}
     \begin{subfigure}[b]{0.16\textwidth}
         \centering
         \includegraphics[width=\textwidth]{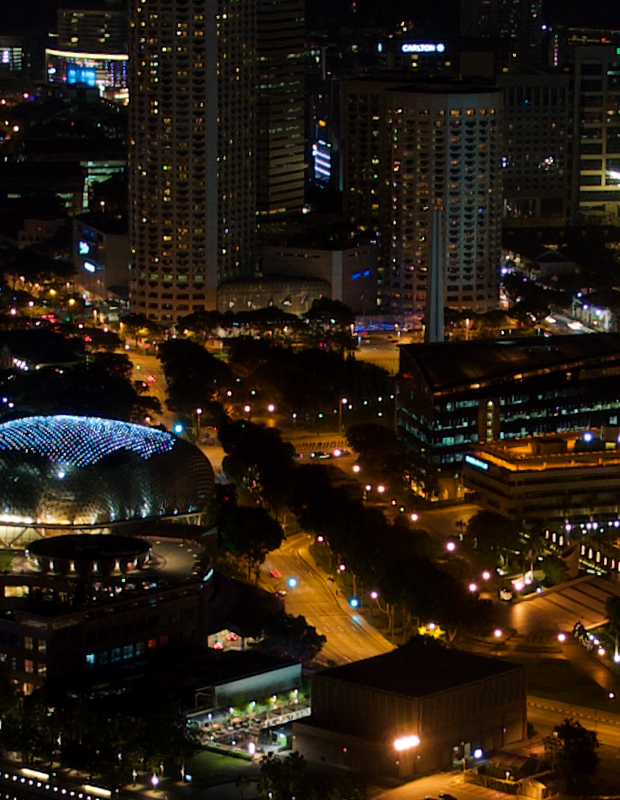}
         \caption{SRC 00007} 
     \end{subfigure}
      \begin{subfigure}[b]{0.16\textwidth}
         \centering
         \includegraphics[width=\textwidth]{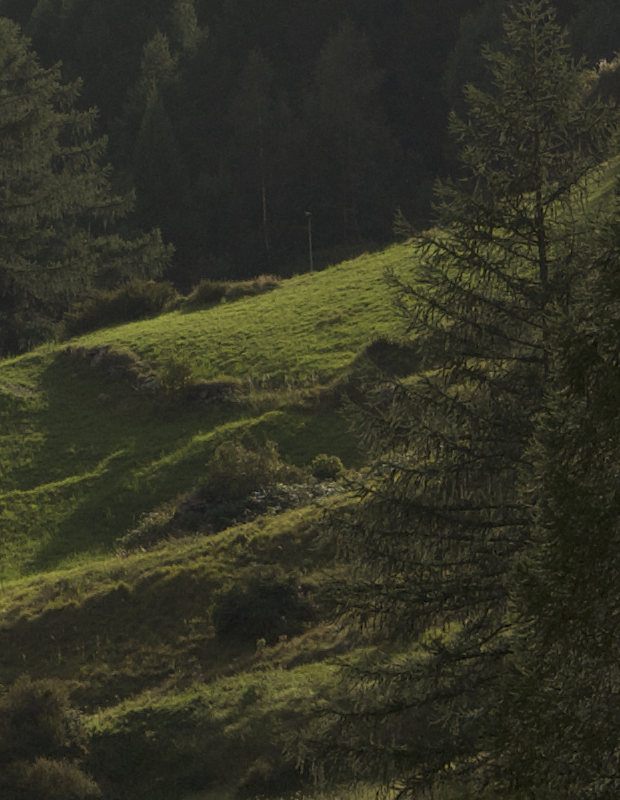}
         \caption{SRC 00009} 
     \end{subfigure}
     \begin{subfigure}[b]{0.16\textwidth}
         \centering
         \includegraphics[width=\textwidth]{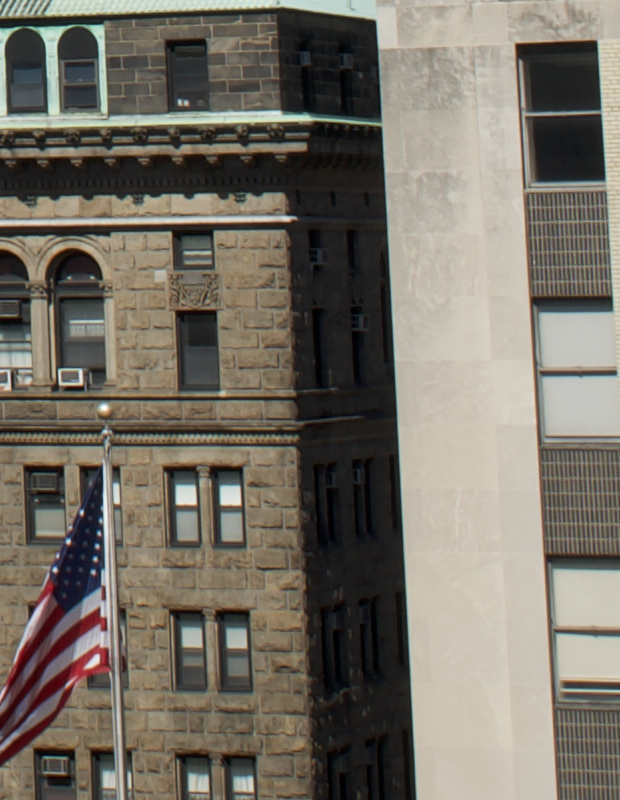}
         \caption{SRC 00010} 
     \end{subfigure}
    \caption{Crops of the reference images adopted in the experiment.}
    \label{fig_dataset}
\end{figure*}

\SubSection{Test material}
Five images from the JPEG AIC-3 dataset were selected to represent a diverse range of image types and content and cropped to a size of $620\times800$ pixels, as presented in Figure ~\ref{fig_dataset}. These images had been compressed with five codes, JPEG, JPEG 2000, VVC Intra, JPEG XL, and AVIF at 10 bitrates each, corresponding approximately to JND values equally spaced from 0.25 to 2.5, as determined by a pairwise comparison experiment~\cite{testolina2023jpeg}.

For the BTC experiment, all 10 distortion levels were considered, while for PTC, only distortion levels numbered 0, 2, 4, 6, 8, and 10 were considered, where 0 indicates the best visual quality (the source image) and level 10 corresponds to compressed images with the strongest artifacts having a perceptual distance from the source of approximately 2.5 JND.

\SubSection{Triplet comparisons and generation of batches }
Four types of triplet questions were used in the experiment:

\begin{itemize}
    \item {Same-codec comparisons}: Triplet questions comparing decoded images with different distortion levels but generated by the same codec.
    \item {Cross-codec comparisons}: Triplet questions comparing decoded images generated by two different codecs, used to help align the scales between images compressed by different codecs.
    \item {Bias-checking comparisons}: Forced choice experiments are often biased since the temporal or spatial ordering of the alternatives for the response has a significant influence on the result. The most direct test to check for such an ordering bias in our experiment is by same-codec triplet questions that compare two identical decoded images.
    \item {Trap questions}: Triplets where one side (left or right) contains a decoded image with the strongest distortion level 10, while the other side displays the original image (level 0). The quality difference between the two images in these trap triplets is intentionally clear, and they are used to identify unreliable workers and determine whether to accept or reject assignments in the crowdsourcing study.
\end{itemize}

All possible same-codec comparisons were included, i.e., 110 and 30 per source and codec for BTC and PTC experiments, respectively. Cross-codec comparisons made up 20\% of same-codec comparisons, resulting in one cross-codec comparison for every five same-codec comparisons. The cross-codec comparisons were chosen randomly targeting similar bitrates. The following table lists the total number of different types of questions used for boosted and plain triplet comparisons. 

\begin{center}
\small
 \begin{tabular}{c c c c c c c}
 & Same-codec & Cross-codec & Bias-checking & Trap questions & Total & Batch size  \\ \midrule[0.1em]
BTC & 2750 & 550 & 100 & 200 & 3600  & 360  \\
PTC & \phantom{0}750 & 150 &  \phantom{0}50 & 100 & 1050 & 105
\end{tabular}
\end{center}

For the BTC and the PTC experiment, the questions were randomly divided into 10 batches as follows. The bias-checking comparisons and the trap questions were uniformly distributed into the batches. The same- and cross-codec questions were joined before splitting them uniformly.

\SubSection{Crowdsourcing Study}

Test subjects were recruited through Amazon Mechanical Turk (MTurk). The BTC and PTC experiments were conducted separately, with several months in between, and 778 and 352 assignments were made available for BTC and PTC, respectively. In each campaign, a crowd-worker could take only a single assignment containing one or two different batches of triplet questions. The batches were selected randomly from the available pool of batches. The sequence of questions within each batch was randomized for each worker. Ethical approval for the experimental procedures and protocols was obtained from the Institutional Review Board of the University of Konstanz. 

\Section{Data analysis}
The data analysis proceeded in four steps:
\setlength{\leftmargin}{\labelwidth}
\begin{enumerate}
    \item \textbf{Filtering of reliable batches.}  The reliability of each batch was evaluated based on same-codec triplet questions, where one test image was the source (distortion level 0), and the other had the highest distortion (level 10). These questions included all trap questions and some same-codec study questions. A batch was deemed reliable if at least 70\% of the responses were correct. Figures \ref{fig_histbtc} and \ref{fig_histptc} show the histograms of the accuracy of batches for BTC and PTC. 
\begin{center}
\small
\begin{tabular}{c c c c c}
    & \multicolumn{2}{c}{Raw data} & \multicolumn{2}{c}{After filtering} \\
    & Batches & Subjects & Batches & Subjects  \\ \midrule[0.1em]
BTC & 1166 & 778 & 615 & 423   \\
PTC & 494 & 352 & 260 & 208 
\end{tabular}
\end{center}
Figure~\ref{fig_bias} shows the responses collected for the bias questions in which there is no difference between the compared stimuli. Clearly, there is a pronounced order bias towards the response ``Right''. After filtering out the unreliable batches, the number of responses ``Left'' and ``Right'' are equal. This indicates that the original order bias is due to unreliable subjects.

\begin{figure*}[p]
    \centering
    \begin{subfigure}[b]{0.3\textwidth}
         \centering
         \includegraphics[trim={0.9cm 0 0cm 0},clip,height=4cm]{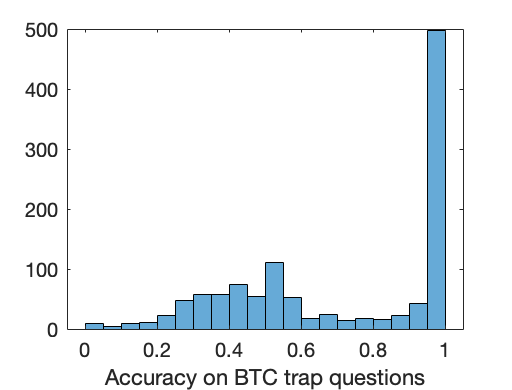}
         \caption{BTC batches}
         \label{fig_histbtc}
     \end{subfigure}
     \begin{subfigure}[b]{0.3\textwidth}
         \centering
         \includegraphics[trim={0.9cm 0 0cm 0},clip,height=4cm]{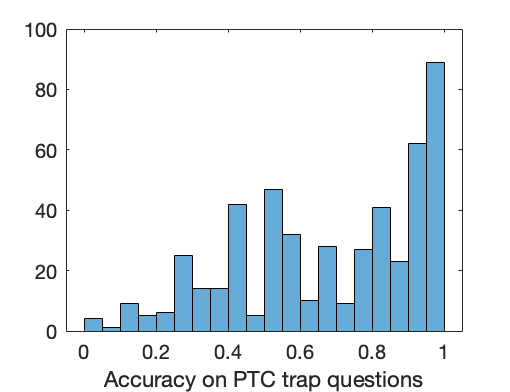}
         \caption{PTC batches}
         \label{fig_histptc}
     \end{subfigure}
     \begin{subfigure}[b]{0.3\textwidth}
         \centering
         \includegraphics[trim={0cm 0 0cm 0},clip,height=3.8cm]{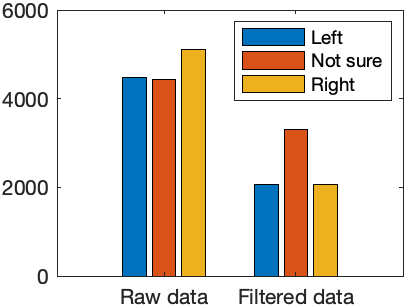}
         \caption{Bias questions}
         \label{fig_bias}
     \end{subfigure}
    \caption{Accuracy of batches for BTC and PTC. Responses for bias questions (right).\\
}
    \label{fig_accuracy}
\end{figure*}

\begin{figure}[p]
    \centering
    \includegraphics[trim={6.9cm 4.5cm 6cm 4.5cm},clip,width=\linewidth]{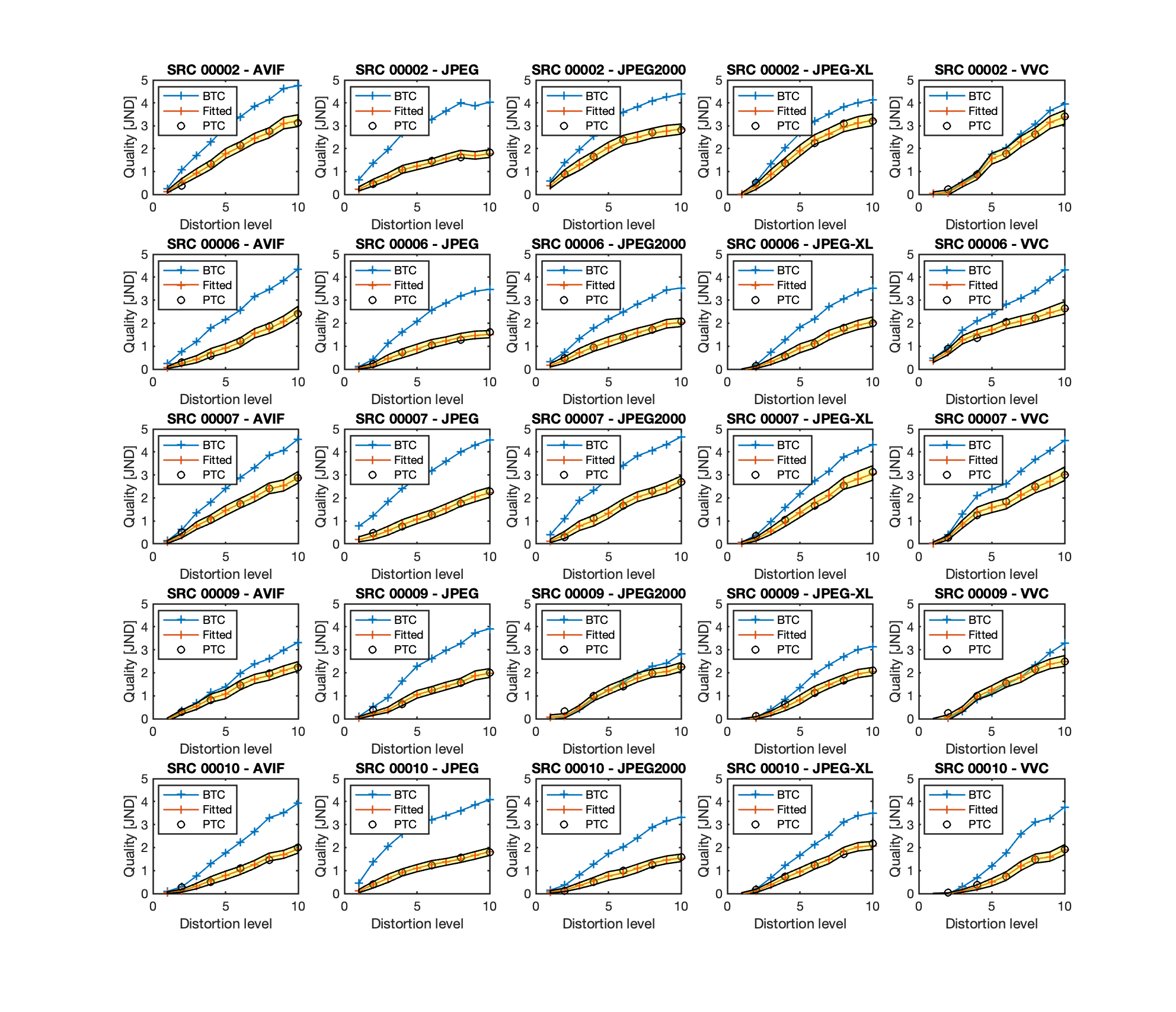}
    \caption{Results of impairment scale reconstruction, aligned boosted scales, and their 95\% confidence intervals. Each row shows the plots for one of the five sources, and columns correspond to the five codecs. The yellow shaded area is narrow and indicates the confidence interval of the aligned scale values.}
    \label{fig_results}
\end{figure}

    \item \textbf{Reconstruction of scales.} The BTC responses were used to reconstruct (boosted) scale values for the corresponding compressed images from the five codecs for each of the five source images. For this purpose, each ``not sure'' response was split into half ``left'' and half ``right''. All responses were interpreted in the sense of two-alternative forced choice in pair comparisons. Then, the reconstruction proceeded by maximum likelihood estimation (MLE) of the scale values in the corresponding Thurstonian Case V model, see \cite[Sect.\ IV.B]{men2021subjective} for more details and a comparison with other approaches. The scale values from the PTC responses were computed in the same way. Note that for Thurstonian reconstruction, it is usually assumed that the variance of the perceived difference in stimuli quality is 1 when the two qualities are 1 unit apart. In order to convert this scale into JND units, one divides the scale values by $\Phi^{-1}(0.75) \approx 0.6745$, where $\Phi$ is the normal cumulative distribution function.
    \item \textbf{Alignment of boosted scales.} To rescale the boosted scores to the range of perceptual quality without boosting, linear regression was used. In this regression, the JND scale values from BTC are the predictor variables, and those from PTC are the target variables for a fit by the degree-two polynomial $ax+bx^2$.
    \item \textbf{Bootstrap for confidence intervals.} $n = 10,000$ bootstrap samples of the (filtered) BTC and PTC data were generated by resampling with the replacement of the responses for each triplet question. The following scale reconstructions gave $n$ values for each scale variable, which yielded the 95\% confidence intervals.
\end{enumerate}

\begin{figure}
    \centering
    \includegraphics[width=0.6\linewidth]{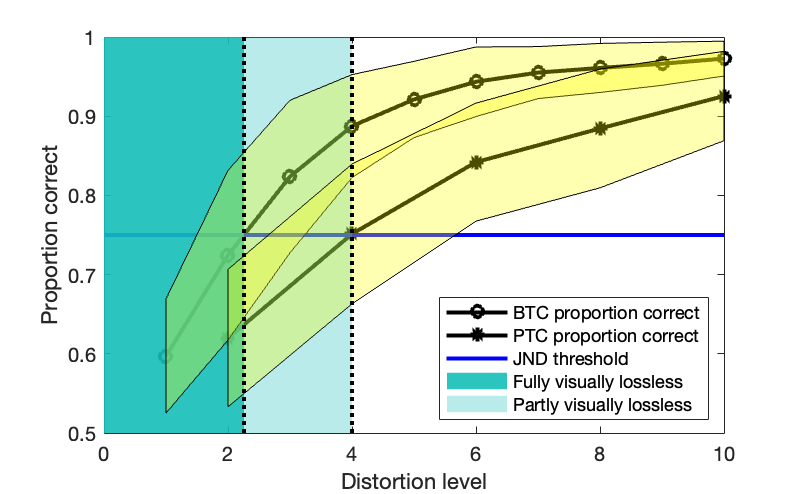}
    \caption{Psychometric functions and JND threshold determination with and without the artifact boosting. The ratio of correct responses to same-codec triplet questions comparing distortion levels 0 and 10 averaged over all sources and codecs. The JND threshold, according to the AIC-2 flicker test, is between the two dotted lines. The dark-shaded region, therefore, is in the visually lossless region, and the lightly shaded region is partly visually lossless.
    }
    \label{fig_baseline}
\end{figure}

%
\Section{Experimental results and discussion}
The alignment of boosted scales can be performed using different procedures. The transformation can be made by one unique polynomial for all sources and codecs or by five polynomials, one per source or codec, respectively. Or one can define a different transformation for each of the 25 source-codec pairs. The latter method yielded the best fitting w.r.t.\ the PTC data, however, at the expense of a higher number of parameters (300 versus 252 or 260 for the other settings). For the performance comparison, we therefore applied the Akaike information criterion that takes the number of parameters into account besides the log-likelihood of the fitted models. The result is that the best fit by taking a different polynomial for each source-codec pair is not outweighed by the cost of the larger number of parameters. Therefore, the option of one transformation per source-codec pair was selected and it was concluded that the boosting transformation modeled by quadratic polynomials depends on the source image as well as on the distortion type.
\newpage

Figure \ref{fig_results} shows an overview of the reconstruction and regression results for all sources and codecs. Generally, the PTC scales (open circles) are within the confidence intervals of the aligned boosted scale values. This shows that the assessment of the visual quality of images with controlled boosting by zooming, artifact amplification, and flicker can successfully be rescaled to match the scales obtained by assessing image quality without boosting. Moreover, the boosted scales are, as expected, larger than the unboosted ones by a factor of about 2. This means that the precision of the aligned scales is also about twice as good as the precision obtainable without boosting. 

Figure~\ref{fig_baseline} presents the psychometric functions of the proportions of correct responses for BTC and PTC that illustrate the JND thresholds with and without boosting. For each tested distortion level, it shows the ratio of correct responses to those same-codec triplet questions in which a compressed image was compared with the source at level 0, which was averaged over all sources and codecs. For example, for PTC at distortion level 4 on the right dotted line, the ratio is 0.75. By definition, this corresponds to a perceived impairment magnitude of 1 JND. Thus, this result confirms that, for distortion level 4, the corresponding decoded images were indeed chosen as anticipated, with a distance of 0.25 JND per level. As expected, the JND threshold for the more sensitive BTC methodology is closer to distortion level 2, indicated by the left dotted line. The threshold for visually lossless compression as defined by JPEG AIC-2 for the flicker test should be between the two dotted lines in the light green-shaded area, since this test is more sensitive than PTC that has a limited manual toggle in place of flicker, and less sensitive than BTC that exploits additional boosting techniques besides flicker.

\Section{Conclusions and future work}

A comprehensive dataset and framework for full-reference subjective quality assessment of high-fidelity decoded images was presented, using boosted and plain triplet comparisons. Boosting techniques are employed to enhance sensitivity to compression artifacts, thereby improving the perceptual ranking of test images. A smaller set of plain triplet comparisons is used in a regression to rescale the boosted quality scales to match those of the original stimuli. The results, provided in JND units, offer new useful information in applications that are not available from traditional DMOS values. For example, this facilitates the estimation of satisfied user ratios in the most relevant range from high to lossless visual quality. The dataset and code will be made available online.
JPEG AIC will continue to develop the proposed framework and the methods for data analysis aiming at an international ISO standard on subjective quality assessment in the high fidelity range.

\Section{Acknowledgments}
 This research is funded by the DFG (German Research Foundation) -- Project ID 496858717 and DFG Project ID 251654672 – TRR 161.  EPFL-affiliated authors acknowledge support from the Swiss National Scientific Research project under grant number 200020\_207918. Instituto Superior Técnico-affiliated authors were supported by FCT/MCTES through national funds under the project DARING with reference PTDC/EEI-COM/7775/2020.

\setlength{\parsep}{0.5em}

\Section{References}
\bibliographystyle{IEEEbib}
\bibliography{dcc}

\end{document}